\documentclass[conference]{IEEEtran}
\IEEEoverridecommandlockouts
\usepackage{cite}
\usepackage{amsmath,amssymb,amsfonts}
\usepackage{algorithmic}
\usepackage{graphicx}
\usepackage{textcomp}
\usepackage{xcolor}
\usepackage{xcolor}
\usepackage{color,soul}

\def\BibTeX{{\rm B\kern-.05em{\sc i\kern-.025em b}\kern-.08em
    T\kern-.1667em\lower.7ex\hbox{E}\kern-.125emX}}
\begin{document}

\title{Efficient Split Learning LSTM Models for FPGA-based Edge IoT Devices\\
\thanks{This work has received funding from the Horizon 2020 research and innovation staff exchange grant agreement No 101086387, and from the  Science Fund of the Republic of Serbia, grant number 6707, REmote WAter quality monitoRing anD IntelliGence – REWARDING}
}

\author{\IEEEauthorblockN{Romina Soledad Molina,\IEEEauthorrefmark{1} Vukan Ninkovic,\IEEEauthorrefmark{2}\IEEEauthorrefmark{3} Dejan Vukobratovic,\IEEEauthorrefmark{2} Maria Liz Crespo,\IEEEauthorrefmark{1} and Marco Zennaro\IEEEauthorrefmark{1}
\IEEEauthorblockA{\IEEEauthorrefmark{1}The Abdus Salam International Centre for Theoretical Physics, Trieste, Italy}
\IEEEauthorblockA{\IEEEauthorrefmark{2}Faculty of Technical Sciences, University of Novi Sad, 
Novi Sad, Serbia}
\IEEEauthorblockA{\IEEEauthorrefmark{3}The Institute for Artificial Intelligence Research and Development of Serbia, 
Novi Sad, Serbia}
}}

\maketitle

\begin{abstract}
Split Learning (SL) recently emerged as an efficient paradigm for distributed Machine Learning (ML) suitable for the Internet Of Things (IoT)-Cloud systems. However, deploying SL on resource-constrained edge IoT platforms poses a significant challenge in terms of balancing the model performance against the processing, memory, and energy resources. In this work, we present a practical study of deploying SL framework on a real-world Field-Programmable Gate Array (FPGA)-based edge IoT platform. We address the SL framework applied to a time-series processing model based on Recurrent Neural Networks (RNNs). Set in the context of river water quality monitoring and using real-world data, we train, optimize, and deploy a Long Short-Term Memory (LSTM) model on a given edge IoT FPGA platform in different SL configurations. Our results demonstrate the importance of aligning design choices with specific application requirements, whether it is maximizing speed, minimizing power, or optimizing for resource constraints.

\end{abstract}


\section{Introduction}


The rapid advancements in Internet of Things (IoT) and cloud computing have transformed environmental monitoring systems, enabling the integration of IoT devices, cloud platforms, and advanced Machine Learning (ML) techniques for efficient data collection, processing, and analysis \cite{mois_2017}. Among ML models, Recurrent Neural Networks (RNNs) excel in analyzing time-series data by capturing temporal dependencies \cite{Husken_2003}. This capability makes them essential for applications such as water quality monitoring, where detecting dynamic patterns in environmental parameters is critical for timely decision-making \cite{bertels2023}. However, deploying RNN models on resource-constrained IoT edge devices poses challenges owing to the limited computational power and energy resources, further compounded by the need for real-time processing in applications requiring rapid responses \cite{gao2019}. Bridging this gap demands innovative strategies that optimize the model performance while respecting the limitations of IoT devices and ensuring data privacy.

Split Learning (SL) has emerged as an innovative approach to address these challenges by partitioning neural networks between edge devices and centralized servers, effectively distributing computational workloads \cite{gupta_2018}. This paper introduces a comprehensive study on the deployment of SL-based Long Short-Term Memory (LSTM) models on Field-Programmable Gate Array (FPGA)-based IoT edge platforms, addressing the demands of resource efficiency and predictive accuracy in time-series applications. The proposed approach integrates advanced model compression techniques—pruning, quantization, and knowledge distillation—to tailor LSTM models for low-power IoT devices. Using a real-world dataset from the Danube River, this study demonstrates how this framework achieves predictive accuracy, low latency, and reduced power consumption. These findings underscore the transformative potential of SL in environmental monitoring, offering scalable and efficient solutions for IoT systems that operate under stringent resource constraints. 
Additionally, the obtained results illustrate design trade-offs: (i) high performance that comes with higher resource and power demands, (ii) balanced design that attempts to optimize both performance and resource utilization, and (iii) efficiency-oriented that prioritizes lower resource and power consumption, sometimes at the expense of speed.


\section{Background}

\subsection{Split Learning}
\label{SL}
\begin{figure}
	\centering
	\includegraphics[width=0.9\linewidth]{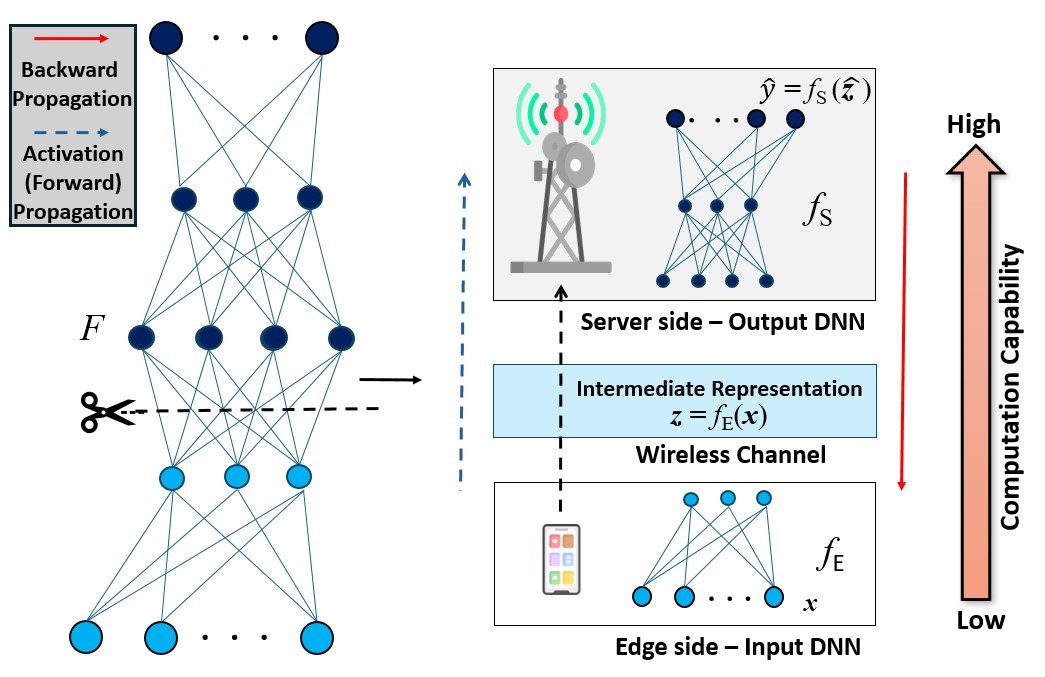}
	\caption{Split learning/inference pipeline with edge and server sub--networks for IoT systems.}
	\label{fig_split}
\end{figure}

Split learning (SL) \cite{gupta_2018} is a distributed machine learning (ML) paradigm designed to alleviate the computational burden on resource-constrained edge devices while maintaining the benefits of collaborative learning. In contrast to traditional centralized or fully distributed models, SL divides the neural network into two sub--networks, with the initial layers processed at the edge (local devices) and the remaining layers processed at the server side, as depicted in Fig. \ref{fig_split}.


During training, an edge device retains data locally, ensuring that raw data are not transmitted across the network \cite{langer_2020}. The neural network is structured as $F=(f_{\text{E}}, f_{\text{S}})$, where $f_{\text{E}}:\mathbb{R}^{N}\rightarrow\mathbb{R}^{M}$ represents the part of the model processed by the edge device, and $f_{\text{S}}:\mathbb{R}^{M}\rightarrow\mathbb{R}$ represents the part executed by the server (Fig. \ref{fig_split}). $N$ and $M$ denote the dimensions of the input and intermediate representation, respectively. We assume $M<N$, i.e., the intermediate representation is considered a compressed version of the input data \cite{shao_2020}.

In both training and inference phases, the edge device sub--network generates an intermediate representation $\boldsymbol{z}=f_{\text{E}}(\boldsymbol{x})$ from the raw input $\boldsymbol{x}$ (Fig. \ref{fig_split}). This intermediate result is transmitted to the server, where the sub--network $f_{\text{S}}$ processes it to produce the final prediction  $\hat{y}=f_{\text{S}}(\boldsymbol{z})$.  In the training phase, the prediction 
$\hat{y}$ is compared with the actual value $y$ and the appropriate loss $\mathcal{L}(\hat{y}, y)$ is calculated \cite{pasquini_2021}. The loss is backpropagated to jointly optimize both the edge and server sub--networks (red solid arrow in Fig. \ref{fig_split}). Additionally, the collaborative nature of the training process facilitates the iterative exchange of intermediate representations and gradients, ensuring data privacy by keeping raw data on the edge \cite{vep_2018}. 

In the context of split inference, this approach is beneficial for IoT applications, where edge devices often have limited memory, processing power, and energy capacity \cite{ayad_2021}. By handling only the initial part of the model, edge devices process raw data and extract low-level features. The intermediate representation $\boldsymbol{z}$, rather than the raw data, is transmitted to the server, which completes the remaining computations and final predictions \cite{vep_2018}. In many cases, the performance of edge devices can be enhanced by deploying Field-Programmable Gate Arrays (FPGAs), offering custom hardware acceleration for the early layers of the model, and providing faster, more responsive, and efficient computation \cite{molanes2018deep}. 


\subsubsection{Split Learning--Based LSTM}
\label{LSTMSPLIT}
This paper builds upon the \textit{LSTMSPLIT} algorithm \cite{jiang_2022}, which follows foundational SL principles and applies them to RNNs (specifically LSTMs). \textit{LSTMSPLIT} efficiently partitions LSTM networks between edge devices and a centralized server, thereby enhancing the processing of sequential data while safeguarding data privacy. This algorithm requires at least two LSTM layers, with the input sequence retained on the edge device. Following the procedure outlined in Section \ref{SL}, the intermediate representation 
$\boldsymbol{z}$ is sent from the edge device's LSTM layer to the server's LSTM layer(s) (dashed blue line, Fig. \ref{fig_split}), while update gradients are sent back to the edge device (solid red line, Fig. \ref{fig_split}), jointly optimizing both sub--networks. Alternatively, integrating SL with RNNs may involve distributing an LSTM layer across multiple edge devices to manage segments within multi-segment training sequences, simultaneously combining split learning and federated learning \cite{abedi_2023}. While this method is intriguing for the problem of interest, it falls beyond the scope of this work.

\subsection{RNN Deployment FPGA Acceleration: State of the Art}

The implementation of RNNs on FPGAs has recently become a prominent research area. This is primarily because FPGA architectures consist of highly reconfigurable circuits, making them ideal for accelerating applications that demand high parallelism, superior performance, and energy efficiency.

Authors in \cite{sun2018} accelerate LSTM for test flight using FPGA devices, demonstrating a speed-up of 28.76x compared with a CPU implementation. The software model is composed of one LSTM and one dense layer. Nevertheless, there is no hardware implementation of the LSTM model, and only the synthesis results are obtained using High-Level Synthesis (HLS). Ribes et al. \cite{ribes2020mapping} perform the deployment of multiple parallel LSTM models on an MPSoC based on FPGA, employing the Roofline model for attainable performance estimation in the exploration of the design space for the hardware accelerator. 
 
Liu et al. \cite{liu2019cloud} propose the design and implementation of a cloud server-oriented FPGA accelerator for LSTM, performing improvements in the communication efficiency between the host server and the FPGA. Pacini et al. \cite{pacini2023} present the extension of a toolflow by adding RNN deployment on FPGA. The authors focus on the quantization strategy for model compression, aiming to speed up the computation of on-board satellite applications. 

He et al. \cite{he2021fpga} focus their research on developing hardware accelerators for LSTM networks using fixed-point arithmetic, systolic arrays for matrix multiplication, and lookup tables for nonlinear functions, demonstrating their effectiveness in image captioning. Bertel et al. \cite{bertels2023} propose integer-only hardware accelerators for RNNs, reducing fixed-point operations but limiting applicability to networks with equal layer widths. Structured pruning for LSTM on high-performance FPGAs, discussed in \cite{wang2019_pruning}, achieved a 7.82× speedup while reducing parameters to 1/8.

Finally, the work in \cite{whisnant2020split} explores the split inference of a CNN targeting FPGA platforms, considering the potential use of this technology in sensor networks. However, only the synthesis results are reported; therefore, there is a lack of details regarding the hardware implementation.

LSTMs exhibit feedback dependence, meaning that each timestep in the sequence relies on the output from the previous one. This sequential nature limits the potential for high levels of parallelism, at which general-purpose processors such as GPUs typically excel.  GPUs are optimized for parallel processing, but the inherent sequential data dependencies in LSTMs effectively reduce their ability to exploit parallelism \cite{he2021fpga}. Consequently, achieving a high performance on general-purpose hardware for LSTM computations can be challenging.  

It can be concluded that the literature reflects a broader trend of increasing FPGA use for accelerating RNN models, minimizing size, and power consumption while maintaining performance and flexibility. This is mainly achieved by increasing the computation-to-communication ratio and reducing the computation costs and memory accesses. Furthermore, as observed in the state-of-the-art, machine learning compression techniques primarily rely on pruning and quantization for LSTM hardware acceleration. 

In contrast to the existing literature, our research employs an ensemble of compression techniques, specifically for model reduction and subsequent implementation on a low-end FPGA, leveraging SL for LSTM. Although the approach in \cite{whisnant2020split} aligns closely with ours, it lacks both FPGA implementation and strategic selection of compression techniques, which are the focal points of our work. 

\section{IoT-Based Environmental Monitoring System}
\subsection{System Model}

We consider a widespread scenario of IoT-based environmental monitoring, focusing on a running example of water quality monitoring of the Danube River near Novi Sad (Serbia). Heterogeneous edge devices with varying processing capabilities continuously sample water quality data and process the data locally before transmitting the intermediate results to a centralized server for further processing. Although many of these devices are low-cost with reduced computational power, the need for minimized latency in communication remains crucial, particularly in cases of environmental accidents, where rapid detection and response are essential to mitigate environmental risks. This scenario highlights the need for compact machine learning models that can efficiently operate on such devices, ensuring timely data processing and decision-making without overwhelming the system's constrained resources.

\begin{figure}
	\centering
	\includegraphics[width=0.83\linewidth]{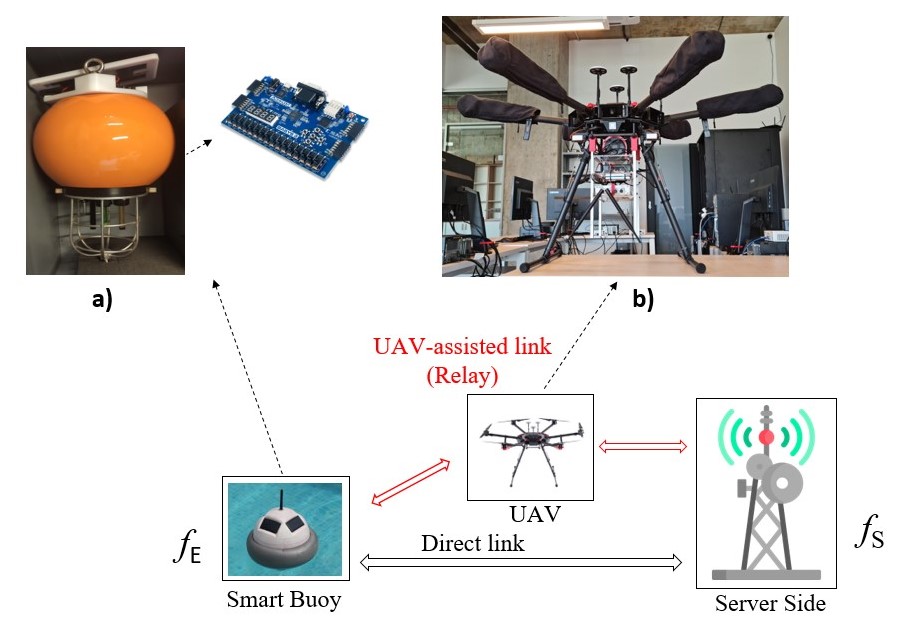}
  \vspace{-3mm}
	\caption[figure caption]{ IoT-based water quality monitoring system with a) smart buoy with FPGA and b) UAV-integrated communication equipment\footnotemark[1]}
	\label{fig_dunav_scen}
\end{figure}

A conventional IoT-based system for water quality monitoring, designed to meet the above requirements, is deployed as illustrated in Fig. \ref{fig_dunav_scen}. The system features a smart buoy equipped with an FPGA and communication chip (Fig. \ref{fig_dunav_scen}a) that facilitates communication with the server. The setup has been thoroughly tested in laboratory environments to ensure its readiness for real-world deployment. Although beyond the scope of this paper, initial tests involving a UAV as a relay link (Fig. \ref{fig_dunav_scen}b) have also been conducted, as detailed in \cite{ninkovic_2024_bcom}. 

\footnotetext[1]{UAV-assisted relay link is out of scope of this work; the initial setup is evaluated in \cite{ninkovic_2024_bcom}.}

Time series environmental data are processed by deploying an LSTM network. This network is split between the smart buoy and the server ($f_{\text{E}}$ and $f_{\text{S}}$ at Fig. \ref{fig_split}), according to the principles outlined in Section \ref{LSTMSPLIT}, enabling efficient analysis of temporal patterns inherent in the collected data. The raw data $\boldsymbol{x}$, representing water quality measurements, undergoes preprocessing at the FPGA deployed at the buoy, producing an intermediate representation $\boldsymbol{z}=f_{\text{E}}(\boldsymbol{x})$. Focusing on the efficient FPGA implementation of RNN networks, we assume ideal channel conditions between the buoy and server; nonetheless, in our previous work, we considered a communication--aware design for the SL paradigm tailored for heterogeneous IoT networks \cite{ninkovic_2024_IoT}. The intermediate representation $\boldsymbol{z}$ is sent to the server, where the prediction $\hat{y}=f_{\text{S}}(\boldsymbol{z})$ is generated. In the offline training phase, the loss is computed based on $\hat{y}$, and $f_{\text{E}}$ and $f_{\text{S}}$ are jointly optimized using algorithms such as stochastic gradient descent (SGD) or its variations (e.g., Adam \cite{adam}).

\subsection{RNN Compression and FPGA Deployment for Time Series Processing}
\label{RNN-FPGA}

 In the context of FPGA platforms, low-end devices are more affordable and are often used for less computationally intensive tasks or applications, where cost and power efficiency are critical. They possess limited memory resources, smaller logic capacity, lower clock speeds, and smaller form factors, but are generally designed to be more power-efficient, making them suitable for battery-powered or low-power embedded systems where minimizing energy consumption is critical.   
 
 These devices enable IoT systems to process data directly on nodes; however, their limited resources require efficient hardware use to avoid compute-bound performance from arithmetic operations. In addition, inefficient data transmission can create memory-bound performance, which is limited by the communication link. To address these challenges, we propose optimizations in both computation -- reducing the memory footprint and arithmetic operations through compression -- and communication, leveraging split inference and LSTM designs that optimize data movement between layers. 

Compression methods \cite{Weng2021} have been employed to deploy DNN algorithms on resource-constrained devices across various fields \cite{Jian2021, Wang2021, CHUN-HSIAN2021, Ngadiuba_2021}. Among them are pruning (P), quantization (Q), and knowledge distillation (KD) \cite{Ganesh_2021}. Q minimizes the memory footprint by reducing the precision of weights and biases, whereas P decreases the number of parameters by eliminating unnecessary neurons and connections. KD \cite{HintonVD15} transfers knowledge from a larger teacher network to a smaller and more efficient student network, enabling it to mimic the teacher's behavior while being computationally lighter. This study combines these methods to leverage their distinct advantages in the compression process.

It is worth noting that integrating KD in the compression pipeline facilitates the deployment of RNN-based models on both low- and high-end FPGA devices. For low-end devices, inference occurs at the edge, while high-end devices enable multiple instances of inference with low latency, as several channels process information simultaneously. The reduced network maintains a good balance between the performance metrics and memory footprint, resulting in models suitable for various technologies. Additionally, network optimization improves the energy efficiency \cite{novac2021quantization} by reducing the high data transfer between the on-chip and off-chip memory.

\begin{figure}[!h]
\begin{center}
\includegraphics[width=0.43\textwidth]{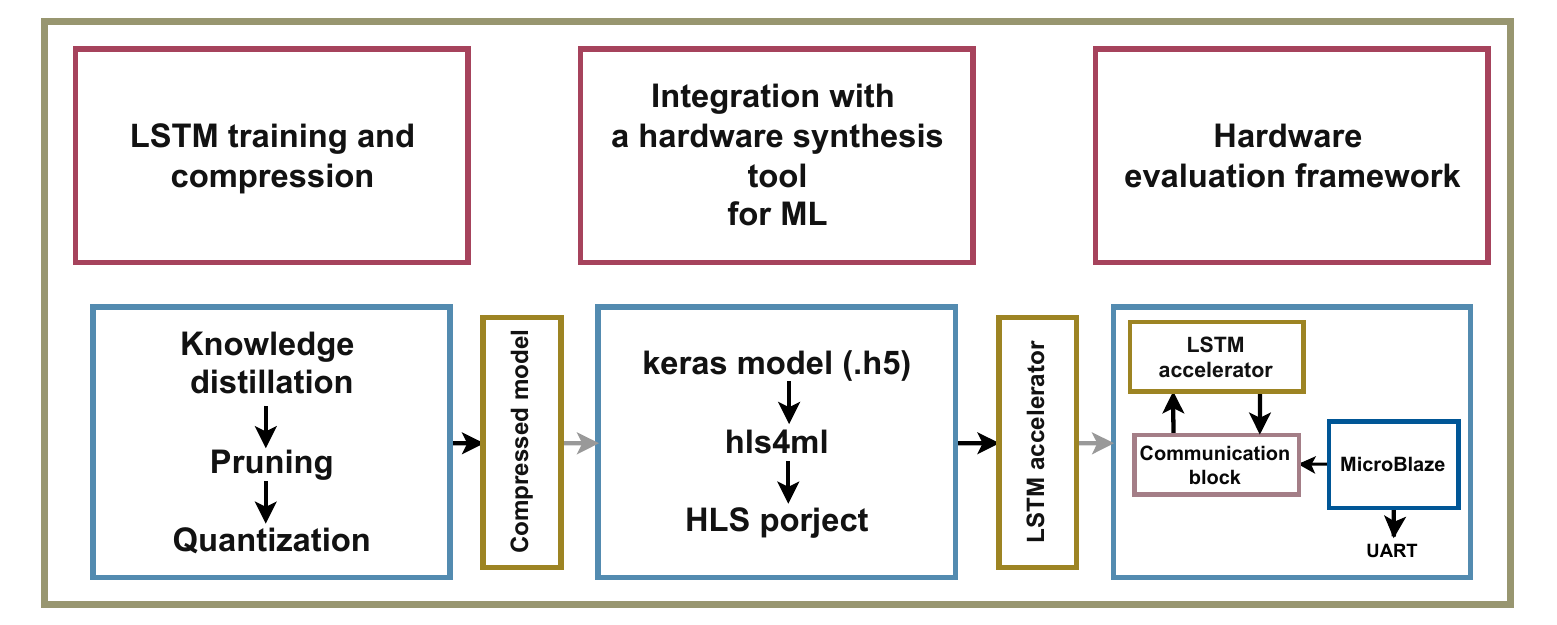}
  \caption{Workflow for model compression and FPGA deployment.}
  \label{FIG:workflow}
\end{center}
\end{figure}

In general, split inference involves partitioning the model, executing part on a resource-constrained device based on FPGA and the remainder in the server. In this study, for the FPGA deployment, we adapted the workflow proposed in \cite{molina2024} to obtain a proper implementation of the RNN, as depicted in Fig. \ref{FIG:workflow}.

As for the employed methodology, an RNN based on LSTM is trained as the teacher model to perform the forecasting task. After the teacher has undergone training, the student architecture is designed to reduce the data transfer of parameters between the node and the server. This involves heuristically  tuning the hyperparameters of the hidden layer to minimize communication overhead, as higher values result in a larger number of intermediate results to transmit. Additionally, a deeper architecture is preferred for this process over a single-layer architecture.

The learning approach for the student network is based on KD. In this study, the distillation loss ($\mathcal{L_{\text{KD}}}$) is obtained according to Eq. \ref{EQ:loss}, replacing the Kullback-Leibler divergence by the mean squared error (MSE) \cite{kim2021comparing}:
\begin{equation}
    \label{EQ:loss}
    \mathcal{L_{\text{KD}}} = \alpha \times \text{MSE}_{\text{TL}} + (1 -\alpha) \times \text{MSE}_{\text{TS}}
\end{equation}
$\mathcal{L_{\text{KD}}}$ is a combination of the MSE with the true labels and MSE between teacher's soft predictions and student's predictions, $\text{MSE}_{\text{TL}}$ and $\text{MSE}_{\text{TS}}$, respectively. The parameter $\alpha$ determines the weight assigned to the distillation loss. A value $\alpha=1$ indicates that only the distillation loss is taken into account, while $\alpha=0$ signifies that the distillation loss is entirely disregarded. $N$ represents the number of data points. $\text{MSE}_{\text{TL}}$ defined as $ \text{MSE}_{\text{TL}} = \frac{1}{N}\sum_{i=1}^{n} (y_i - \hat{y}_i)^2$, allows obtaining the MSE considering the true labels and the student's predictions ($y_i$ and $\hat{y}_i$, respectively). The MSE between teacher's soft predictions ($w_i$) and student's predictions ($\hat{y}_i$) is defined according to $\text{MSE}_{\text{TS}} =  \frac{1}{N}\sum_{i=1}^{n} (w_i - \hat{y}_i)^2$.

After the distillation process, post-pruning and post-quantization are performed to reduce the memory footprint of the FPGA. Next, a High-Level Synthesis (HLS) project is created using hls4ml \cite{Duarte_2018} to generate a hardware description of the forecasting stage. The LSTM model is then split, with a portion synthesized for deployment on the FPGA, while the remaining architecture runs outside the edge device. Finally, a hardware evaluation framework based on a soft-core processor is developed to assess the LSTM accelerator on the FPGA.

\subsection{System overview}

\label{SECTION:sysOverview}
\begin{figure}[!h]
\begin{center}
\includegraphics[width=0.4\textwidth]{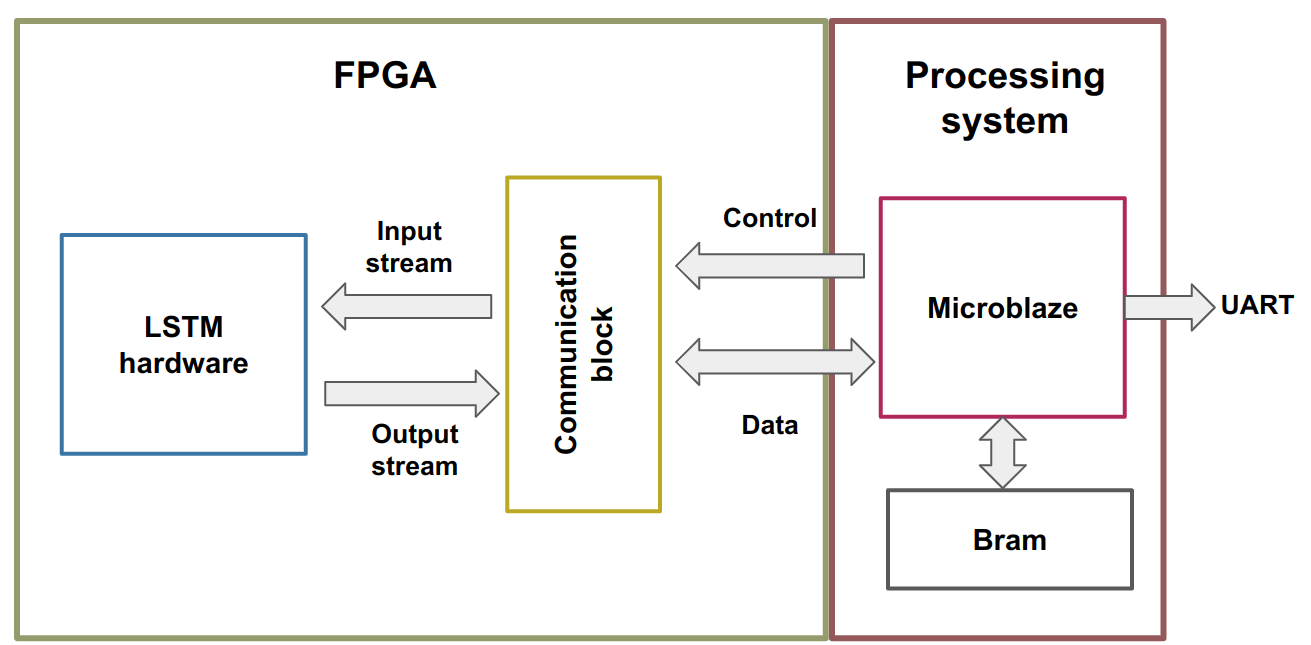}
  \caption{Hardware evaluation framework.}
  \label{FIG:BD_HW}
\end{center}
\end{figure}

The performance of the LSTM accelerator is evaluated using the hardware assessment framework illustrated in Fig. \ref{FIG:BD_HW}. A soft-core processor functions as a control system, sending data and control signals to the FPGA. The LSTM inference hardware generated from the HLS is integrated into the framework design, allowing seamless interchangeability during the testing phases. Communication between the processor and the inference hardware is facilitated by ComBlock \cite{ComBlockSPL2019}, which streamlines the interaction between the soft-core processor and FPGA. The input data stream is fed into the LSTM core, and upon completing the computation, the output  
is returned to the processor. The evaluation step can be monitored through UART.

\section{Performance Evaluation}
\subsection{Dataset Generation}

The dataset utilized in this study for IoT-based water quality monitoring provides a comprehensive foundation for assessing river health and identifying potential pollution. Collected from the Danube River near Novi Sad, it contains 3,264 instances of daily measurements recorded between November 2013 and October 2022, of which 70\%\ are allocated for training and 30\%\ for testing purposes. This dataset includes seven key water quality parameters: temperature, pH, electrical conductivity, dissolved oxygen, oxygen saturation, ammonium, and nitrite. Although our setup currently supports multivariate prediction, we focus specifically on tracking dissolved oxygen levels over several days, storing these values for temporal analysis and LSTM preprocessing. This choice aligns with our primary goal of optimizing the model for low-power, resource-constrained devices.  

\subsection{Training, compression, and hardware generation}
To forecast dissolved oxygen levels in the Danube River, we use two distinct LSTM architectures: teacher (LSTM-DO-T) and student (LSTM-DO-S). In both cases, the input size is set to 15, meaning that the last 15 dissolved oxygen measurements are used to predict the next value. 

The teacher architecture LSTM-DO-T is a sequential model beginning with an LSTM layer, followed by a fully connected (FC) layer, another LSTM layer, and ending with a final FC layer to produce the output. 
Regarding the student architecture, LSTM-DO-S consists of four layers organized as two LSTM layers, followed by two FC layers. The hyperparameters for both architectures during the learning process are presented in Table I.

\begin{table}[!h]
\renewcommand{\arraystretch}{1.3}
\label{TABLE:hyperparameters}
  \caption{List of hyperparameters for LSTM-DO-T and LSTM-DO-S. }
\centering
\resizebox{0.43\textwidth}{!}{
\begin{tabular}{c|cc}
    \hline
\textbf{Hyperparameter} &  LSTM-DO-T & LSTM-DO-S  \\
\hline
\textbf{Type of Layers} & LSTM-FC-LSTM-FC & LSTM-LSTM-FC-FC  \\
\textbf{Batch size} & 8 & 16  \\
\textbf{Epochs} & 32 & 32   \\
\textbf{Optimizer} & Adam & Adam  \\
\textbf{Learning rate} & 0.001 & 0.001    \\
\textbf{Regularization} & L2 (0.001) & L2 (0.01)    \\
\textbf{Loss function} & $\mathcal{L_{\text{MSE}}}$ & $\mathcal{L_{\text{KD}}}$ ($\alpha$=0.1) \\
\textbf{Regression metrics} & MSE, MAE &  MSE, MAE \\ 
  \hline
\end{tabular}
}
\end{table}

Once the student network is generated, post-pruning is applied with a target sparsity of 70\% and post-quantization with 8-bit fixed-point precision to reduce the memory footprint and computational complexity. After completing training and compression, the student network layers are deployed on the FPGA, with the LSTM model’s weights and biases stored in the FPGA's on-chip memory. To facilitate this deployment, hls4ml is used to generate the HLS project.  

SL allows the computation to start or end at any layer, enabling the selection of an optimal split point. Using the HLS tool, the architecture is partitioned to enhance hardware efficiency by implementing specific layers on the FPGA. This study focuses on the hardware acceleration of LSTM layers and examines their impact on FPGA deployment. To this end, we construct three LSTM inference variations: 

\begin{itemize}
    \item LSTM-DO-S: the complete student network,
    \item Split-A: a configuration with two stacked LSTM layers,
    \item Split-B: a configuration containing a single LSTM layer.
\end{itemize}

 After generating the hardware accelerators, evaluations are conducted using the framework described in Section \ref{SECTION:sysOverview}.
 
\subsection{Results} 
\textit{Experimental setup:} FPGA assessment was performed on the Basys 3 development board based on Artix-7™ (part: xc7a35tcsg325-1) FPGA from ADM/Xilinx. Power consumption was measured using an Innovateking-EU digital multimeter. For the performance evaluation of the model, the mean absolute error (MAE), mean squared error (MSE), and coefficient of determination ($R^2$) were used. 




Hardware metrics including resource utilization, latency, and scalability are reported for the LSTM-based hardware accelerators. Resource usage is provided in terms of Block RAM (BRAM), Look-Up-Table (LUT), Flip Flop (FF), and Digital Signal Processing (DSP) elements. The latency is computed in $\mu s$ considering a clock of 80 $\text{MHz}$. The scalability of the system is obtained through  $\text{SC} = \min \bigg\{ \frac{\text{BRAM\_{T}}}{\text{BRAM\_{PE}}}, \frac{\text{DSP\_{T}}}{\text{DSP\_{PE}}}, \frac{\text{LUT\_{T}}}{\text{LUT\_{PE}}}, \frac{\text{FF\_{T}}}{\text{FF\_{PE}}} \bigg\}$, where $\text{BRAM\_{T}}$, $\text{DSP\_{T}}$, $\text{LUT\_{T}}$, $\text{FF\_{T}}$ represent the total amounts of each hardware resource and $\text{BRAM\_{PE}}$, $\text{DSP\_{PE}}$, $\text{LUT\_{PE}}$, $\text{FF\_{PE}}$ define the number of resources used by the processing element (PE), that in this study corresponds to the LSTM-based hardware. 

  Table \ref{TABLE:LSTM_teacher_student} presents the performance assessment of the teacher and student models in terms of parameters, size, MAE, MSE, and $R^2$. A slight performance degradation owing to compression is observed, although it remains acceptable for this type of implementation. The compression ratio observed is 45.90$\times$ in terms of KB, making the LSTM-DO-S model suitable for resource-constrained devices. 

\begin{table}[h]
\renewcommand{\arraystretch}{1.3}
  \caption{Performance evaluation of the teacher (LSTM-DO-T) and student (LSTM-DO-S) models. }
    \label{TABLE:LSTM_teacher_student}
\centering
\resizebox{0.45\textwidth}{!}{
\begin{tabular}{c|ccccccc}
    \hline
 \textbf{Model} &  \textbf{Parameters} &\textbf{Size [KB]} & \textbf{MAE} &\textbf{MSE} &  \textbf{$R^2$}    \\    \hline 
LSTM-DO-T  & 39,951 & 156.06 & 0.0520  &0.0081 & 0.97 \\  
LSTM-DO-S & 871 & 3.40 & 0.0574 & 0.0087  & 0.96 \\ 
 \hline 
\end{tabular}
}
\end{table}

Table III presents the FPGA implementation metrics for the LSTM-DO inference, including the full student network (LSTM-DO-S), split configuration (Split-A) with two stacked LSTM layers on the FPGA, and split configuration (Split-B) with one LSTM layer on the FPGA. The results show that Split-B offers the best efficiency in terms of overall resource utilization and is the most power efficient design. However, Split-B trades latency for improved modularity and potential pipelining, whereas Split-A and LSTM-DO-S prioritize the speed. These results highlight the design trade-offs: (i) high performance with LSTM-DO-S, (ii) a balanced approach with Split-A, and (iii) an efficiency-oriented focus with Split-B. 

\begin{table}[!h]
\renewcommand{\arraystretch}{1.2}
\label{TABLE:implementationMetrics}
  \caption{Implementation results for LSTM-based accelerators. FPGA part: xc7a35tcsg325-1.}
\centering
\resizebox{0.43\textwidth}{!}{
\begin{tabular}{c|ccc}
    \hline
\textbf{Metrics} &  LSTM-DO-S & Split-A   & Split-B  \\
\hline
\textbf{BRAM} & 31\% & 16\%  & 20\% \\
\textbf{DSP} & 55\% & 33\%  & 17.7\% \\
\textbf{LUT} & 50\% & 40\%  & 28\%  \\
\textbf{FF} & 17\% & 16\%   &  13\%  \\ 
\hline
\textbf{Overall max utilization} & 61\% & 60\%   & 48\% \\ 
\textbf{Latency} [$\mu s$] & 2.82 &  2.85   &  3.54 \\ 
\textbf{Frequency [$\text{MHz}$]} & 80 & 80     &  80 \\ 
\textbf{Scalability} & 1 & 1     &  2 \\ 
\textbf{Power consumption [W]} & 0.887 & 0.862  & 0.829\\ 

  \hline
\end{tabular}
}
\end{table}

A significant distinction between Split-A and Split-B is the size of the intermediate outputs generated by the LSTM architecture -- 5 and 150, respectively, -- which affects latency due to data transfers between BRAM and the inference core.   
 
In an IoT setting, where multiple measurements are simultaneously captured through sensors, hardware resource utilization analysis reveals that Split-B achieves more efficient resource use (LUT: 28\%). By contrast, the split-A and LSTM-DO-S configurations can instantiate only a single processing element on the FPGA (DSP: 55\% and LUT: 40\%, respectively). The overall FPGA maximum resource utilization accounts for the associated hardware, including the MicroBlaze soft-core processor, BRAM, communication blocks, inference core, and additional IP cores for reset and interconnections. Because our setup supports multivariate prediction, the Split-B strategy enables the implementation of two LSTM-based accelerators, thereby allowing parallel ML computations on a resource-constrained device.


\section{Conclusion}

 From the results obtained in this paper, one can easily recognize the potential of SL in resource-limited hardware platforms. A key design aspect is the strategy for partitioning the architecture and the amount of data that needs to be transferred between the memory and the processing block. Multiple ML model hardware instances can be instantiated in the FPGA part, enabling computation parallelization and addressing the case when input signals come from different sensors. Furthermore, the combination of KD, P, and Q offers faster forecasting and improved memory efficiency, while maintaining performance. The resulting RNN/LSTM models are more suitable for deployment in production environments on resource-limited edge IoT nodes for local data pre-processing.

\bibliographystyle{IEEEtran}

\bibliography{conference_101719}

\end{document}